# Solving Linear Equations by Classical Jacobi-SR Based Hybrid Evolutionary Algorithm with Uniform Adaptation Technique

A. R. M. Jalal Uddin Jamali*, *M. M. A. Hashem**, *M. Mahfuz Hasan**  and *Md. Bazlar Rahman*.

*Dept. of Mathematics
**Dept. of Computer Science and Engineering
Khulna University of Engineering and Technology
Khulna 9203, Bangladesh

## Abstract

Solving a set of simultaneous linear equations is probably the most important topic in numerical methods. For solving linear equations, iterative methods are preferred over the direct methods specially when the coefficient matrix is sparse. The rate of convergence of iteration method is increased by using Successive Relaxation (SR) technique. But SR technique is very much sensitive to relaxation factor, $\omega$. Recently, hybridization of classical Gauss-Seidel based successive relaxation technique with evolutionary computation techniques have successfully been used to solve large set of linear equations in which relaxation factors are self-adapted. In this paper, a new hybrid algorithm is proposed in which uniform adaptive evolutionary computation techniques and classical Jacobi based SR technique are used instead of classical Gauss-Seidel based SR technique. The proposed Jacobi-SR based uniform adaptive hybrid algorithm, inherently, can be implemented in parallel processing environment efficiently. Whereas Gauss-Seidel-SR based hybrid algorithms cannot be implemented in parallel computing environment efficiently. The convergence theorem and adaptation theorem of the proposed algorithm are proved theoretically. And the performance of the proposed Jacobi-SR based uniform adaptive hybrid evolutionary algorithm is compared with Gauss-Seidel-SR based uniform adaptive hybrid evolutionary algorithm as well as with both classical Jacobi-SR method and Gauss-Seidel-SR method in the experimental domain. The proposed Jacobi-SR based hybrid algorithm outperforms the Gauss-Seidel-SR based hybrid algorithm as well as both classical Jacobi-SR method and Gauss-Seidel-SR method in terms of convergence speed and effectiveness.

KewWord: Jacobi-SR Method, Gauss-Seidel-SR Method, Evolutionary Algorithm, Adaptive Technique.

## 1. Introduction

Solving a large set of simultaneous linear equations is probably the most important topic in numerical methods. Systems of linear equations are associated with many problems in engineering and science, as well as with applications of mathematics to the social sciences and the quantitative study of business, statistics and economic problems. After invent of easily accessible computers, it is possible and practical for us to solving large set of simultaneous linear algebraic equations. Now for appropriate decision of the physical problems, it is sometimes desired an appropriate algorithm which converged rapidly for solving physical problems. For example, short-term weather forecast, image processing, simulation to predict aerodynamics performance which of these applications involve the solution of very large set of simultaneous equations by numerical methods and time is an important factor for practical application of the results. If the algorithm of solving equations can be implemented in parallel processing environment efficiently, it can easily decrease a significance time to get the result.

There are mainly two classical numerical methods- direct method and iterative method to solve systems of linear equations. For large set of linear equations, especially for sparse and structured coefficient (matrices) equations, iterative methods are preferable as iterative method are unaffected by round off errors to a large extent [1]. The well-known classical numerical iterative methods are the Jacobi method and Gauss-Seidel method. The rate of convergence, as very slow for both cases, can be accelerated by using Successive Relaxation (SR) technique [2]. But the speed of convergence depends on relaxation factor (denoted by $\omega$) with a necessary condition for the convergence is $0 < \omega < 2$ and SR technique is very much sensitive to relaxation factor [3, 4]. However, it is often very difficult to estimate the optimal relaxation factor, which is a key parameter of the SR technique [5, 6].

On the other hand the Evolutionary Algorithms (EA) are stochastic algorithms whose search methods model some natural phenomena: genetic inheritance and Darwinian strife for survival [7, 8, 9]. Almost all of the works on EA can be classified as evolutionary optimization (either numerical or combinatorial) or evolutionary learning. But Fogel and Atmar [10] used linear equation solving as test problems for comparing recombination, inversion operations and Gaussian mutation in an evolutionary algorithm. A linear system of the form

$$b_i = \sum_{j=1}^{n} a_{ij} x_j, \qquad i = 1, 2, \cdots, n$$

was used in their study. The worth of an individual that encoded ($x_1, x_2, \ldots, x_n$) was defined according to the error function

$$E = \sum_{i=1}^{n} E_i$$

where

$$E_i = |\sum_{j=1}^{n} a_{ij} x_j - b_i |, \qquad i = 1, 2, \cdots, n$$

However, the emphasis of their study was not on equation solving, but rather on comparing the effectiveness of recombination relative to mutation. No comparison with classical equation-solving methods was given, and only small problems ($n = 10$) were considered [10].

Recently for solving linear equations, Jun et. al. [11] have proposed a hybrid evolutionary algorithms which is developed by integrating classical Gauss-Seidel based SR method and evolutionary computation techniques with uniform self adaptation technique. Jamali et. al. [12] have also recently have developed a hybrid evolutionary algorithms by integrating classical Gauss-Seidel based SR method and evolutionary computation techniques with Time variant self adaptation technique. The idea of self-adaptation was also applied in many different fields [12,13, 14, 15, 16]. In both of the above discussed hybrid algorithms, the classical Gauss-Seidel based SR method was used. But as Gauss-Seidel based SR method cannot be implemented in parallel processing environment efficiently, so above discussed both Gauss-Seidel based hybrid evolutionary cannot be implemented, inherently, in parallel processing environment efficiently. On the other hand, Jacobi based SR method can be implemented in parallel processing environment efficiently [1]. So Jacobi-SR based hybrid evolutionary algorithm can be implemented in parallel processing environment efficiently. Note that in hybrid evolutionary algorithm, individuals of population can be implemented in parallel processing environment explicitly. But as we have discussed above that if the algorithms can be implemented in parallel processing environment, it can easily decrease a significance time to get the result. So for trying to eliminate above-mentioned problems and to decrease the time of convergence (by using parallel processors), in this paper, a new hybrid evolutionary algorithm is proposed in which evolutionary computation techniques are used with classical Jacobi-SR method. The uniform adaptation (UA) technique is used for adaptation of relaxation factors. The proposed Jacobi-SR based hybrid algorithm does not require a user to guess or estimate the optimal relaxation factor $\omega$. The proposed algorithm initializes relaxation factors stochastically in a given uniformly distributed domain and "evolves" it. The proposed algorithm integrates the classical Jacobi-based SR method with evolutionary computation techniques, which uses initialization, recombination, mutation, adaptation, and selection mechanisms. It makes better use of a population by employing different equation-solving strategies for different individuals in the population. Then these individuals can exchange information through recombination and the error is minimized by mutation and selection mechanisms. The effectiveness of the proposed hybrid algorithm is compared with that of classical Jacobi based SR method, Gauss-Seidel based SR method and Gauss-Seidel based Uniform adaptive GSBUA hybrid algorithm in experimental domain. Also for the validity of the proposed algorithm, two theorems are stated and are proved. The preliminary investigation has showed that proposed algorithm outperforms the classical numerical methods; proposed algorithm is comparable with GSBUA hybrid algorithm in sequential processing environment. Another significant property of this proposed algorithm is that this algorithm inherently can be implemented in parallel processing environment efficiently. And the preliminary investigation has also showed that, in parallel processing environment, proposed algorithm converges a fraction of time compared to GSBUA algorithm.

## 2. The Basic Equations of Classical Jacobi Based SR Method

The system of $m$ linear equations with $n$ unknown $x_1, x_2, \cdots, x_n$ can be written as

$$\sum_{j=1}^{n} a_{ij} x_j = b_i, \quad (i = 1, 2, \cdots, m) \tag{1}$$

or equivalently, with $m = n$, in matrix form

$$\mathbf{Ax} = \mathbf{b} \tag{2}$$

where

$$\mathbf{A} = (a_{ik}) = \begin{bmatrix} a_{11} & a_{12} & \cdots & a_{1n} \\ a_{21} & a_{22} & \cdots & a_{2n} \\ \vdots & \vdots & \ddots & \vdots \\ a_{m1} & a_{m2} & \cdots & a_{nn} \end{bmatrix}, \quad \mathbf{x} = \begin{bmatrix} x_1 \\ x_2 \\ \vdots \\ x_n \end{bmatrix} \text{ and } \mathbf{b} = \begin{bmatrix} b_1 \\ b_2 \\ \vdots \\ b_n \end{bmatrix}$$

such that $\mathbf{A} \in \Re^n \times \Re^n$, $\mathbf{x} \in \Re^n$ and $\mathbf{b} \in \Re^n$, here $\Re$ is real number.

Note that for unique solution $|\mathbf{A}| \neq 0$. Assume without loss of generality that none of the diagonal entries is of zero; otherwise interchange it rows. And decomposed the coefficient matrix $\mathbf{A}$ as

$$\mathbf{A} = (\mathbf{D} + \mathbf{U} + \mathbf{L})$$

where $\mathbf{D} = (d_{ij})$ is a diagonal matrix, $\mathbf{L} = (l_{ij})$ is a strictly lower triangular matrix and $\mathbf{U} = (u_{ij})$ is a strictly upper triangular matrix. Then Eq. (2) can be rewrite as

$$(\mathbf{D} + \mathbf{U} + \mathbf{L})\mathbf{x} = \mathbf{b},$$

or $\quad \mathbf{D}\mathbf{x} = \mathbf{b} - (\mathbf{U} + \mathbf{L})\mathbf{x}$

or $\quad \mathbf{x} = \mathbf{D}^{-1}\mathbf{b} - \mathbf{D}^{-1}(\mathbf{U} + \mathbf{L})\mathbf{x}$

or $\quad \mathbf{x} = \mathbf{H}_j \mathbf{x} + \mathbf{V}_j$ $\hfill$ (3)

where $\mathbf{H}_j = \mathbf{D}^{-1}(-\mathbf{L} - \mathbf{U})$, is called Jacobi iteration matrix, and $\mathbf{V}_j = \mathbf{D}^{-1}\mathbf{b}$.

Now express Eq. (3) in component-wise, and then an equivalent form for the system is

$$x_i = -\sum_{\substack{k=1 \\ k \neq i}}^{n} \frac{a_{ik}}{a_{ii}} x_k + \frac{b_i}{a_{ii}}, \quad i = 1, \cdots, n \hfill (4)$$

And construct the sequence $\{\mathbf{x}^{(k)}\}$ for an initial vector $\mathbf{x}^{(0)}$ by setting

$$\begin{cases} \mathbf{x}^{(k+1)} = \mathbf{H}_j \mathbf{x}^{(k)} + \mathbf{V}_j \text{ with} \\ \mathbf{x}^{(k)} = \left(x_1^{(k)}, x_2^{(k)}, \cdots, x_n^{(k)}\right)^t \end{cases} \quad \text{for } k = 0, 1, 2, \cdots. \hfill (5)$$

Which is called Jacobi iteration method or simply Jacobi method. Now express Eq. (5) in component-wise, then Jacobi method becomes

$$x_i^{(k+1)} = \frac{b_i}{a_{ii}} - \sum_{\substack{j=1 \\ j \neq i}}^{n} \frac{a_{ij}}{a_{ii}} x_j^{(k)}, \quad i = 1, \cdots, n \text{ and } k = 0, 1, 2, \cdots \hfill (6)$$

Again by using SR technique [1, 2] in Eq. (3), then the system of equations can be expressed as (in component-wised)

$$x_i^{(k+1)} = x_i^{(k)} + \frac{\omega}{a_{ii}} \left( b_i - \sum_{j=1}^{n} a_{ij} x_j^{(k+1)} \right) \hfill (7)$$

Which is called Jacobi based Successive Relaxation (SR) method. And in matrix form, Eqn.(7) can be rewrite as

$$\mathbf{x}^{(k+1)} = \mathbf{H}_\omega \mathbf{x}^{(k)} + \mathbf{V}_\omega \hfill (8)$$

where

$$\mathbf{H}_\omega = \mathbf{D}^{-1}\{(1-\omega)\mathbf{I} - \omega(\mathbf{L} + \mathbf{U})\} \hfill (9)$$

and $\mathbf{V}_\omega = \omega \mathbf{D}^{-1}\mathbf{b}$;

Here $\mathbf{H}_\omega$ is called Jacobi-SR iteration matrix, $\mathbf{I}$ is called identity matrix, $\omega \in (\omega_L, \omega_U)$ is called relaxation factor which influence the convergence rate of the method greatly; also $\omega_L$ and $\omega_U$ are denoted as lower and upper boundary values of $\omega$.

## 3. The Proposed Hybrid Algorithm

The key idea behind the proposed hybrid algorithm is to self-adapt the relaxation factor used in classical Jacobi based SR method. Similar to many other evolutionary algorithms, the proposed hybrid algorithm also always maintains a population of approximate solutions to the system of linear equations. Each solution is represented by an individual. In this proposed hybrid algorithm, different relaxation factors are used to solve equations. And each relaxation factor is associated with each individual. The relaxation factors are adapted based on the fitness of individuals. The fitness of an individual is evaluated based on the error of an approximate solution. For example, given an approximate solution (i.e., an individual) $\mathbf{x}$, its error is defined by $\|e(\mathbf{x})\| = \|\mathbf{A}\mathbf{x} - \mathbf{b}\|$, where $\|.\|$ is called norm of the vector. The relaxation factor is adapted after each generation, depending on how well an individual performs (in term of error).

The initial population is generated randomly from the field $\Re^n$. Different individuals use different relaxation factors. Recombination in the hybrid algorithm involves all individuals in a population. If the population size is $N$, then the recombination will have $N$ parents and generates $N$ offspring through linear combination. Mutation is achieved by performing one iteration of Jacobi based SR method as given by Eq. (8). The mutation is stochastic since $\omega$ used in the iteration is initially generated randomly between $\omega_L$ and $\omega_U$ and $\omega$ is adapted stochastically in each generation (iteration). The main steps of the Jacobi-SR Based Uniform Adaptive (JBUA) hybrid evolutionary algorithm described as follows:

**Step-1 Initialization**

An initial population $\mathbf{X}^{(0)}$, of size $N$, is randomly generated from $\Re$ for approximate solution to the system of linear equations Eq. (2) as

$$\mathbf{X}^{(0)} = \{\mathbf{x}_1^{(0)}, \mathbf{x}_2^{(0)} \ldots \mathbf{x}_N^{(0)}\} \qquad (10)$$

Also generate randomly corresponding initial relaxation factors, $\omega_i$, for each individual within boundary ($\omega_L, \omega_U$). Each individual $\mathbf{x}_i \in \Re^n$ and superscripts (.) of the Eq. (10) denotes number of generation (iteration number). So the population of $k$-th generation (iteration) is given by

$$\mathbf{X}^{(k)} = \{\mathbf{x}_1^k, \mathbf{x}_2^k, \ldots, \mathbf{x}_N^k\} \qquad (11)$$

**Step-2 Recombination**

Then generate an intermediate recombination population $\mathbf{X}^{(k,c)}$ as

$$\mathbf{X}^{(k,c)} = \mathbf{R}\mathbf{X}^{(k)} \qquad (12)$$

where $\mathbf{X}^{(k,c)} = \begin{pmatrix} \mathbf{x}_1^{(k,c)} \\ \mathbf{x}_2^{(k,c)} \\ \vdots \\ \mathbf{x}_N^{(k,c)} \end{pmatrix}$, $\mathbf{X}^{(k)} = \begin{pmatrix} \mathbf{x}_1^{(k)} \\ \mathbf{x}_2^{(k)} \\ \vdots \\ \mathbf{x}_N^{(k)} \end{pmatrix}$ and $\mathbf{R} = (r_{ij})_{N \times N}$

Here $\mathbf{R}$ is stochastic matrix [17], and superscript ($k, c$) denotes $k$-th generation with crossover so that $\mathbf{x}_i^{(k,c)}$ denotes $i$-th crossover-individual in $k$-th generation.

**Step 3. Mutation**

Generate the next intermediate population $\mathbf{X}^{(k,m)}$ from $\mathbf{X}^{(k,c)}$ as follows: For each individual $\mathbf{x}_i^{(k,c)}$ ($1 \leq i \leq N$) in population $\mathbf{X}^{(k,c)}$ produces an offspring according to classical Jacobi-based SR method as follows:

$$\mathbf{x}_i^{(k,m)} = \mathbf{H}_{\omega_i} \mathbf{x}_i^{(k,c)} + \mathbf{V}_{\omega_i}, \quad i = 1, 2, \ldots, N \qquad (13)$$

where superscript ($k, m$) denotes $k$-th generation with mutation so that $\mathbf{x}_i^{(k,m)}$ denotes $i$-th mutated offspring in $k$-th generation and $\omega_i$ denotes relaxation factor associated with the $i$-th individual. Note that only one iteration is carried out for each mutation.

**Step-4 Adaptation**

Calculate the errors of each individual of offspring according to the fitness of each pair of individuals (in this paper ranking the fitness value of all offspring and successively adjacent two of them are compared) the relaxation factors are self-adapted by following formula:

Let $\mathbf{x}^{(k,m)}$ and $\mathbf{y}^{(k,m)}$ be two selected offspring individuals and $\|e(\mathbf{x}^{(k,m)})\|$ and $\|e(\mathbf{y}^{(k,m)})\|$ are the corresponding errors (fitness value). Then the corresponding relaxation factors $\omega_x$ and $\omega_y$ respectively are adapted as follows:

(a) If $\|e(\mathbf{x}^{(k,m)})\| > \|e(\mathbf{y}^{(k,m)})\|$,

   (i) then move $\omega_x$ toward $\omega_y$ by using

$$\omega_x^m = (0.5 + p_x)(\omega_x + \omega_y) \tag{14}$$

   where $p_x$ is a random number in (-0.01, 0.01), and

   (ii) move $\omega_y$ away from $\omega_x$ using $\omega_y^m = \begin{cases} \omega_y + p_y(\omega_U - \omega_y), & \text{when } \omega_y > \omega_x \\ \omega_y + p_y(\omega_L - \omega_y), & \text{when } \omega_y < \omega_x \end{cases}$

   where $p_y$ is a random number in (0.008, 0.012).

(b) If $\|e(\mathbf{x}^{(k,m)})\| < \|e(\mathbf{y}^{(k,m)})\|$, adapt $\omega_x$ and $\omega_y$ in the same way as above but reverse the order of $\omega_x^m$ and $\omega_y^m$.

(c) If $\|e(\mathbf{x}^{(k,m)})\| = \|e(\mathbf{y}^{(k,m)})\|$, no adaptation. So that

$$\omega_x^m = \omega_x \text{ and } \omega_y^m = \omega_y.$$

Note that $\omega_x^m$ and $\omega_y^m$ denote adapted relaxation factors. Also note that the idea of adapting parameters can be applied to many different domains. For example, back-propagation algorithms for neural-network training can be accelerated using self-adaptive learning rate [14].

### Step-5 Selection and Reproduction

The best $N/2$ individuals in offspring population $\mathbf{X}^{(k,m)}$ will reproduce (i.e. each individual generates two offspring), and then form the next generation $\mathbf{X}^{(k+1)}$ of $N$ individuals. Note that according to the selection mechanism used here, we choose $N$ (population size) as an even number.

### Step-6 Halt

If the error of the population $\|e(\mathbf{X})\| = \min\{\|e(\mathbf{z})\|: \mathbf{z} \in \mathbf{X}\}$ is less than a given threshold $\eta$ then the algorithm terminates; otherwise, go to **Step -2**.

## 4. Convergence Theorems

The following theorem establishes the convergence of the hybrid algorithm.

**Theorem-1**: If there exist an $\varepsilon$ $(0 < \varepsilon < 1)$ such that, for the norm of $\mathbf{H}_\omega$,

$$\|\mathbf{H}_\omega\| < \varepsilon < 1,$$

then $\quad \lim_{k \to \infty} \mathbf{x}^{(k)} = \mathbf{x}^*,$

where $\mathbf{x}^*$ is the solution to the linear system of equations i.e.,

$$\mathbf{A}\mathbf{x}^* = \mathbf{b}$$

**Proof**: Let $\mathbf{X}^{(k)}$ be the population at $k$-th iteration so that

$$\mathbf{X}^{(k)} = \{\mathbf{x}_i^{(k)} : i = 1, 2, \cdots, N\}$$

Let $\mathbf{e}_i^{(k)}$ be the error between the approximate solution $\mathbf{x}_i^{(k)}$ and exact solution $\mathbf{x}^*$ i.e.

$$\mathbf{e}_i^{(k)} = \|\mathbf{x}_i^{(k)} - \mathbf{x}^*\|$$

Then according to the recombination

$$\mathbf{x}_i^{(k,c)} = \sum_{j=1}^{N} r_{ij} \mathbf{x}_j^{(k)}, \quad i = 1, 2, \cdots, N$$

$\therefore \quad \|\mathbf{e}_i^{(k,c)}\| = \|\mathbf{x}_i^{(k,c)} - \mathbf{x}^*\|$

$$= \|\sum_{j=1}^{N} r_{ij} \mathbf{x}_j^{(k)} - \mathbf{x}^*\|$$

$$= \|\sum_{j=1}^{N} r_{ij} \mathbf{x}_j^{(k)} - \sum_{j=1}^{N} r_{ij} \mathbf{x}^*\|$$

$\because \quad \sum_{j=1}^{N} r_{ij} = 1$ and $r_{ij} \geq 0, \quad i = 1, 2, \cdots, N$

$\therefore \quad \|\mathbf{e}_i^{(k,c)}\| = \|\sum_{j=1}^{N} r_{ij}(\mathbf{x}_j^{(k)} - \mathbf{x}^*)\|$

$$\leq \sum_{j=1}^{N} \|r_{ij}(\mathbf{x}_j^{(k)} - \mathbf{x}^*)\| \leq \sum_{j=1}^{N} \|r_{ij}\| \|\mathbf{x}_j^{(k)} - \mathbf{x}^*\| \qquad \text{(by Schwarz's Inequality)}$$

$\therefore \quad \|\mathbf{e}_i^{(k,c)}\| \leq \sum_{j=1}^{N} r_{ij} \|\mathbf{x}_j^{(k)} - \mathbf{x}^*\|$

$\therefore \quad \|\mathbf{e}_i^{(k,c)}\| < \max\{\|\mathbf{e}_i^{(k)}\| : j = 1, 2, \cdots, N\}$

Again according to mutation for $i = 1, 2, \cdots, N$

$$\mathbf{x}_i^{(k,m)} = \mathbf{H}_{\omega_i} \mathbf{x}_i^{(k,c)} + \mathbf{V}_{\omega_i}$$

and also since $\mathbf{x}^* = \mathbf{H}_{\omega_i} \mathbf{x}^* + \mathbf{V}_{\omega_i}$ then

$$\mathbf{x}_i^{(k,m)} - \mathbf{x}^* = \mathbf{H}_{\omega_i}(\mathbf{x}_i^{(k,c)} - \mathbf{x}^*)$$

$$\therefore \quad \|\mathbf{x}_i^{(k,m)} - \mathbf{x}^*\| = \|\mathbf{H}_{\omega_i}(\mathbf{x}_i^{(k,c)} - \mathbf{x}^*)\|$$

$$\leq \|\mathbf{H}_{\omega_i}\| \, \|(\mathbf{x}_i^{(k,c)} - \mathbf{x}^*)\|$$

$$\leq \|\mathbf{H}_{\omega_i}\| \, \|\mathbf{e}_i^{(k,c)}\|$$

$$< \|\mathbf{H}_{\omega_i}\| \, \max\{\|\mathbf{e}_i^{(k)}\|; i = 1, 2, \cdots, N\}$$

$$\therefore \quad \|\mathbf{e}_i^{(k,m)}\| < \varepsilon \max\{\|\mathbf{e}_i^{(k)}\|; i = 1, 2, \cdots, N\}$$

Now according to the selection mechanism, we have for $i = 1, 2, \cdots, N$

$$\|\mathbf{e}_i^{(k+1)}\| \leq \|\mathbf{e}_i^{(k,m)}\| < \varepsilon \max\{\|\mathbf{e}_i^{(k)}\|; i = 1, 2, \cdots, N\}$$

This implies that the sequence $\{\max\{\|\mathbf{e}_j^{(k,+1)}\|; j = 1, 2, \cdots, N\}; k = 0, 1, 2, \cdots\cdots\}$ is strictly monotonic decreasing and thus convergent.

The following theorem justifies the adaptation technique for relaxation factors used in proposed hybrid evolutionary algorithms.

**Theorem –2:** *Let $\rho(\omega)$ be the spectral radius of matrix $\mathbf{H}_\omega$, $\omega^*$ be the optimal relaxation factor, and let $\omega_x$ and $\omega_y$ are any two relaxation factors. Assume $\rho(\omega)$ is monotonic decreasing when $\omega < \omega^*$ and $\rho(\omega)$ is monotonic increasing when $\omega > \omega^*$. Also consider $\rho(\omega_x) > \rho(\omega_y)$. Then*

(i) $\rho(\omega_x^m) < \rho(\omega_x)$, *when* $\omega_x^m = (0.5 + \tau)(\omega_x + \omega_y)$

*where* $\tau \in (-E_x, E_x)$, *and* $E_x = \dfrac{\omega_y \sim \omega_x}{2(\omega_x + \omega_y)}$,

(ii) $\rho(\omega_y^m) < \rho(\omega_y)$ *when* $\omega_y^m = \omega_y + \delta \, \text{sign}(\omega_y - \omega_x)$, *where* $0 < \delta \leq |\omega^* - \omega_y|$,

*and* $\omega^* < \omega_y < \omega_x$ *or* $\omega_x < \omega_y < \omega^*$

**Proof**: We first assume that $\rho(\omega)$ is monotonic decreasing when $\omega < \omega^*$ and $\rho(\omega)$ is monotonic increasing when $\omega > \omega^*$. Also let $\omega_x$ and $\omega_y$ are any two-relaxation factors and let $\rho(\omega_x) > \rho(\omega_y)$. Then there will be two cases:

**Case-1** Both $\omega_x, \omega_y < \omega^*$ (see **Fig**. 1):

Since $\rho(\omega)$ is monotonic decreasing when $\omega < \omega^*$ and as assume $\rho(\omega_x) > \rho(\omega_y)$; so $\omega_x < \omega_y$.

Now since $\omega_x^m = \omega_x + (0.5 + \tau)(\omega_x + \omega_y)$ where $\tau \in [-E_x, E_x]$, so $\omega_x^m$ must go away from $\omega_x$ and lies between $\omega_x$ and $\omega_y$; i.e. $\omega_x < \omega_x^m < \omega_y$ (see **Fig**. 1). Now as $\rho(\omega)$ is monotonic decreasing when $\omega < \omega^*$, so $\rho(\omega_y) < \rho(\omega_x^m) < \rho(\omega_x)$

i.e. $\rho(\omega_x^m) < \rho(\omega_x)$ (Proved the part (i))

Again since $\omega_x < \omega_y$ and $\omega_y^m = \omega_y + \delta \, \text{sign}(\omega_y - \omega_x)$ where $0 < \delta \leq |\omega^* - \omega_y|$ and $\omega_x < \omega_y < \omega^*$. So $\text{sign}(\omega_y - \omega_x)$ is positive and therefore $\omega_y^m$ go away from $\omega_x$ and $\omega_y$ to $\omega^*$. That is $\omega_y < \omega_y^m \leq \omega^*$ (see **Fig**.1). Now since $\rho(\omega)$ is monotonic decreasing when $\omega < \omega^*$, therefore $\rho(\omega^*) < \rho(\omega_y^m) < \rho(\omega_y)$

i.e. $\rho(\omega_y^m) < \rho(\omega_y)$. (Proved the part (ii))

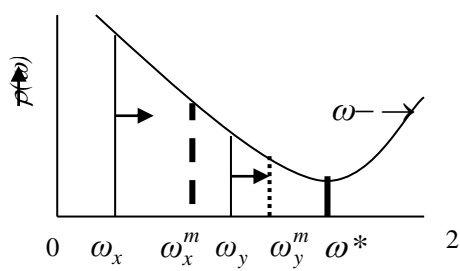 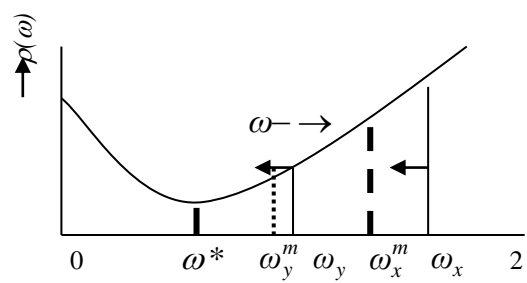

**Figure** 1: Decrease both spectral radii of $\omega_x$, $\omega_y$ when $\omega_x, \omega_y < \omega^*$

**Figure** 2: Decrease both spectral radii of $\omega_x$, $\omega_y$ when $\omega_x, \omega_y > \omega^*$

**Case-2** Both $\omega_x, \omega_y > \omega^*$ (see **Fig**. 2):

Since $\rho(\omega)$ is monotonic increasing when $\omega > \omega^*$ and as assuming $\rho(\omega_x) > \rho(\omega_y)$ so $\omega_x > \omega_y$.

Now since $\omega_x^m = \omega_x + (0.5+\tau)(\omega_x + \omega_y)$ where $\tau \in [-E_x, E_x]$ so $\omega_x^m$ must go away from $\omega_x$ and lies between $\omega_y$ and $\omega_x$; i.e. $\omega_y < \omega_x^m < \omega_x$ (see **Fig**. 2). Now as $\rho(\omega)$ is monotonic increasing when $\omega > \omega^*$, so $\rho(\omega_y) < \rho(\omega_x^m) < \rho(\omega_x)$

i.e. $\rho(\omega_x^m) < \rho(\omega_x)$. (Proved the part (i))

Again since $\omega_y^m = \omega_y + \delta \, \text{sign}(\omega_y - \omega_x)$ where $0 < \delta \leq |\omega^* - \omega_y|$ and $\omega_x < \omega_y < \omega^*$. So $\text{sign}(\omega_y - \omega_x)$ is negative and therefore $\omega_y^m$ go away from $\omega_y$ (as well as $\omega_x$ and ) to $\omega^*$. Therefore we have $\omega^* \leq \omega_y^m < \omega_y$ (see **Fig**.2). Now since $\rho(\omega)$ is monotonic increasing when $\omega > \omega^*$, therefore $\rho(\omega^*) \leq \rho(\omega_y^m) < \rho(\omega_y)$

i.e. $\rho(\omega_y^m) < \rho(\omega_y)$. (Proved the part (ii))

Hence the theorem is proved completely.

## 5. Numerical Experiment

In order to evaluate the effectiveness and efficiency of the proposed JBUA hybrid algorithm, numerical experiments had been carried out on a number of problems to solve the systems of linear Eq. (2) of the form:

$$\mathbf{A}\mathbf{x} = \mathbf{b}$$

The proposed JBUA hybrid algorithm, used in all of our experiments, was very simple and had population size two ($N = 2$). That is, only two individuals were used. The recombination matrix **R** in all through the experiments, was chosen as follows:

Since only two individuals were used in a population of our experiments, if the fitness of the first individual was better then the second (using Eqn. (4.2.7)), then let

$$\begin{pmatrix} \mathbf{x}_1^{(k,c)} \\ \mathbf{x}_2^{(k,c)} \end{pmatrix} = \begin{pmatrix} 1.0 & 0 \\ 0.99 & 0.01 \end{pmatrix} \begin{pmatrix} \mathbf{x}_1^{(k)} \\ \mathbf{x}_2^{(k)} \end{pmatrix}$$

else let (15)

$$\begin{pmatrix} \mathbf{x}_1^{(k,c)} \\ \mathbf{x}_2^{(k,c)} \end{pmatrix} = \begin{pmatrix} 0.01 & 0.99 \\ 1.0 & 0.0 \end{pmatrix} \begin{pmatrix} \mathbf{x}_1^{(k)} \\ \mathbf{x}_2^{(k)} \end{pmatrix}$$

The following settings were also valid all through the experiments:

The dimension of unknown variables was $n = 100$, each individual **x** of population **X** was initialized from the domain $\Re^{100} \in (-30, 30)$ randomly and uniformly. For each experiment, a total of ten independent runs were conducted and the average results are reported here.

Now first problem (say $P_0$) was to solve linear equations, Eq. (2), where the parameters are as follows:

$a_{ii} = 2n$ ; $b_i = i$ for $i = 1, \cdots, n$ and $a_{ij} = j$ for $i \neq j$, $i, j = 1, \cdots, n$.

The problem was to be solved with an error smaller than $\eta = 10^{-12}$.

**Table** 1
Comparison of Jacobi-based SR method and proposed JBUA hybrid algorithm

| Iteration | Jacobi based SR method with initial | | JBUA hybrid algorithm with initial | |
|---|---|---|---|---|
| | $\omega = 0.50$ | $\omega = 1.50$ | $\omega_1 = 0.50$ and | $\omega_2 = 1.50$ |
| 01 | 1.83716e+04 | 2.36221e+04 | 4.3972e+04 | 2.26891e+04 |
| 100 | 3.97643e+03 | 2.05325e+04 | 9.42877e+00 | 3.10284e+00 |
| 200 | 3.14860e+01 | 1.83852e+02 | 3.52673e-03 | 1.02643e-03 |
| 300 | 1.06612e+00 | 4.23743e+01 | 2.78793e-04 | 1.19089e-04 |
| 400 | 4.43217e-02 | 9.54315e+00 | 1.23254e-05 | 1.02244e-06 |
| 500 | 1.04843e-02 | 6.08937e+00 | 1.80543e-06 | 1.03781e-06 |
| 600 | 7.55472e-03 | 4.28310e+00 | 7.15730e-08 | 2.43217e-08 |
| 700 | 2.35390e-03 | 2.61748e+00 | 3.98569e-09 | 1.01475e-09 |
| 800 | 1.02362e-03 | 2.12982e+00 | 2.25191e-10 | 1.03283e-10 |
| 900 | 7.27216e-04 | 1.63231e+00 | 8.44612e-11 | 5.00851e-11 |
| 1000 | 1.32542e-04 | 9.76833e-01 | 6.96453e-12 | 3.74284e-12 |

**Table** 1 and **Table** 2 show the numerical results achieved by the classical Jacobi-based SR method and proposed JBUA hybrid evolutionary algorithm with initial relaxation factors, $\omega$ = 0.5, 1.50 and $\omega$ = -1.0, 1.0 respectively. Four experiments were carried out using classical Jacobi based SR method with relaxation factors $\omega$ = 0.5, 1.50 and $\omega$ = -1.0, 1.0. And Two experiments were carried out using the proposed algorithm, one with initial relaxation factors $\omega_1$ = *0.5 and* $\omega_2$ = 1.5 and the other with initial relaxation factors $\omega_1$ = *-1.0* and $\omega_2$ = 1.0. It is very clear from the **Tables** 1 and 2 that the proposed algorithm performed much better than the classical Jacobi based SR method. Proposed JBUA algorithm with different initial relaxation factors, $\omega_1$ and $\omega_2$, have all found approximate solutions with an error smaller than $\eta = 10^{-12}$ within 1000 generations, while none of the classical Jacobi based SR method could find an approximate solution with an error smaller than $\eta = 10^{-12}$ after 1000 generations, no matter which relaxation factors had been used. After 1000 generations, there was at least eight orders of magnitude difference between the error generated by the classical Jacobi-based SR method and that produced by the proposed JBUA hybrid algorithm.

**Table** 2
Comparison of Jacobi-based SR method and proposed JBUA hybrid algorithm

| Iteration | Jacobi based SR method with initial | | JBUA hybrid algorithm with initial | |
|---|---|---|---|---|
| | $\omega = -1.0$ | $\omega = 1.0$ | $\omega_1 = -1.0$ and | $\omega_2 = 1.0$ |
| 01 | 8.372113e+13 | 1.18508e+04 | 1.372113e+12 | 4.3972e+04 |
| 100 | Diverge | 3.89678e+03 | 5.22573e+2 | 2.36573e+2 |
| 200 | Diverge | 1.26643e+02 | 1.35920e-00 | 1.01321e-00 |
| 300 | Diverge | 1.75359e+01 | 2.19745e-02 | 1.79832e-02 |
| 400 | Diverge | 2.34710e+00 | 5.66802e-04 | 3.23152e-04 |
| 500 | Diverge | 9.83765e-01 | 3.47889e-05 | 1.89475e-05 |
| 600 | Diverge | 3.26554e-01 | 2.22358e-07 | 1.39126e-07 |
| 700 | Diverge | 5.06362e-02 | 5.89688e-09 | 3.26786e-09 |
| 800 | Diverge | 1.03243e-02 | 8.74730e-11 | 4.82132e-11 |
| 900 | Diverge | 8.68931e-03 | 3.57647e-12 | 1.32256e-12 |
| 1000 | Diverge | 1.23574e-03 | 1.23741e-13 | 1.19243e-13 |

**Table** 3
The dynamical change of relaxation factors, $\omega$, for corresponding individuals at different generations for proposed JBUA hybrid algorithm

| Iteration | Value of $\omega$'s for 1st experiment of JBUA | | Value of $\omega$'s for 2nd experiment of | |
|---|---|---|---|---|
| 1 | 0.5 | 1.5 | -1.0 | 1.0 |
| 100 | 1.039819 | 1.05214 | 0.869122 | 0.871368 |
| 200 | 1.08041 | 1.08407 | 0.972992 | 0.97667 |
| 300 | 1.094001 | 1.096638 | 1.039424 | 1.043368 |
| 400 | 1.086654 | 1.098117 | 1.057982 | 1.057956 |
| 500 | 1.072153 | 1.085534 | 1.060547 | 1.059654 |
| 600 | 1.08393 | 1.080362 | 1.072739 | 1.068253 |
| 700 | 1.082872 | 1.088507 | 1.080413 | 1.068221 |
| 800 | 1.07965 | 1.070871 | 1.085379 | 1.093159 |
| 900 | 1.087312 | 1.076113 | 1.089493 | 1.090912 |
| 1000 | 1.051892 | 1.054571 | 1.082053 | 1.098993 |

Besides, **Table** 2 shows that when it was used initial relaxation factor $\omega_1 = -1.0$ and $\omega_2 = 1.0$, the proposed hybrid algorithm adapted relaxation factors and converged rapidly. Whereas in classical Jacobi based SR method, at relaxation factor $\omega = -1.0$, diverged very rapidly.

**Table** 3 shows how relaxation factors are changed dynamically and adapted itself when proposed JBUA hybrid algorithm progressed. Though for the first experiment, initial relaxation factors were $\omega_1 = 0.5$ and $\omega_2 = 1.5$, after 1000 iteration, they became $\omega_1 = 1.051892$ and $\omega_2 = 1.054571$ respectively by using uniform adaptation technique. Again for second experiment, the initial relaxation factors $\omega_1 = -1.0$ and $\omega_2 = 1.0$ again adapted to $\omega_1 = 1.082053$ and $\omega_2 = 1.098993$ respectively. Note that the optimal relaxation of this problem is near to 1.20.

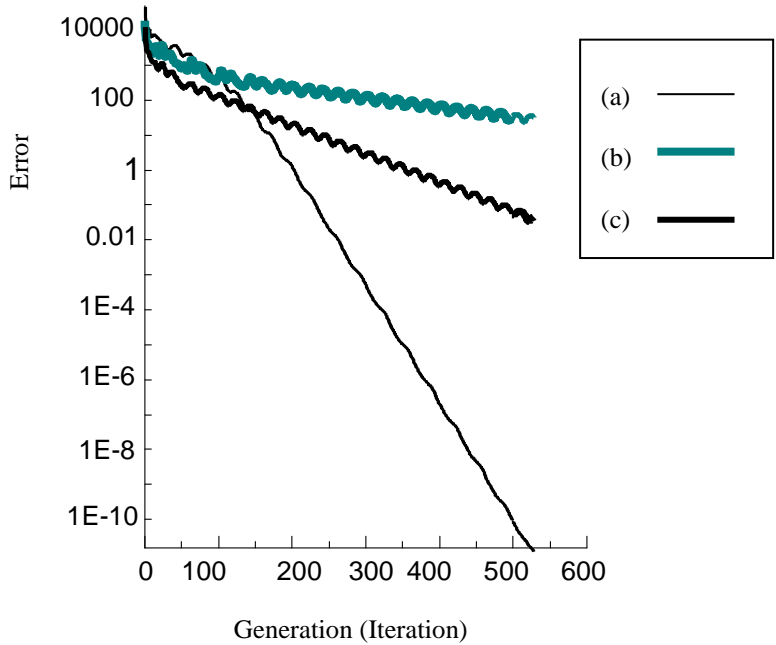

**Figure 3**: Curve (a) represents proposed JBUA hybrid generation history, curve (b) represents classical Jacobi-SR iteration history and curve (c) represents classical Jacobi iteration history.

**Fig.** 3 represents the characteristics the same problem $P_1$, produced by the proposed JBUA hybrid algorithm with initial relaxation factors $\omega_1 = 0.5$ and $\omega_2 = 1.5$, Classical Jacobi based SR method with relaxation factor $\omega = 1.5$ and Classical Jacobi method (i.e. relaxation factor $\omega = 1.0$). It is clear, from **Fig.** 3, that proposed JBUA hybrid algorithm outperforms the classical Jacobi based SR method ($\omega \neq 1.0$) as well as Classical Jacobi method ($\omega = 1.0$). It also shows that classical SR technique is extremely sensitive to the relaxation factor $\omega$.

The nest experiment was to solve the system of linear equations Eq. (2) with parameters given bellows:

$a_{ii} = n^2$ ; $b_i = i$ for $i = 1, \cdots, n$ ; and $a_{ij} = j$ for $i \neq j$, $i, j = 1, \cdots, n$

The problem was to be solved with an error smaller than $\eta = 10^{-6}$.

**Table 4**
Number of generations required by the proposed JBUA hybrid evolutionary algorithm for various randomly generated relaxation factors.

| S.N | Value of initial $\omega$ used in JBUA algorithm | | Generation | S. N | Value of initial $\omega$ used in JBUA algorithm | | Generation |
|---|---|---|---|---|---|---|---|
| | $\omega_1$ | $\omega_2$ | | | $\omega_1$ | $\omega_2$ | |
| 1 | 0.882629 | 0.576721 | 18 | 18 | 0.147400 | 1.293030 | 18 |
| 2 | 0.174561 | 1.066589 | 18 | 19 | 0.466370 | 0.806335 | 18 |
| 3 | 0.767151 | 0.779663 | 19 | 20 | 0.528137 | 0.598145 | 22 |
| 4 | 0.208069 | 1.189331 | 18 | 21 | 0.000612 | 1.99893 | 19 |
| 5 | 0.365723 | 1.445007 | 20 | 22 | 1.455200 | 0.350342 | 20 |
| 6 | 0.781494 | 1.817566 | 19 | 23 | 0.175537 | 1.374817 | 22 |
| 7 | 1.984436 | 0.176941 | 19 | 24 | 0.796021 | 1.254456 | 19 |
| 8 | 0.829712 | 0.614502 | 18 | 25 | 0.972229 | 0.411808 | 17 |
| 9 | 0.816284 | 0.318726 | 19 | 26 | 0.406982 | 1.538879 | 23 |
| 10 | 1.058289 | 0.239319 | 18 | 27 | 0.593445 | 1.769950 | 24 |
| 11 | 0.335449 | 1.771667 | 18 | 28 | 1.380371 | 0.600525 | 18 |
| 12 | 0.889896 | 0.235901 | 17 | 29 | 1.296631 | 0.787231 | 21 |
| 13 | 1.500244 | 0.704773 | 23 | 30 | 0.324280 | 1.209351 | 18 |
| 14 | 0.726257 | 0.590576 | 18 | 31 | 1.228880 | 0.654846 | 22 |
| 15 | 0.296082 | 1.597473 | 22 | 32 | 1.420959 | 0.068787 | 22 |
| 16 | 0.372437 | 1.692566 | 19 | 33 | 1.828491 | 0.482605 | 19 |
| 17 | 1.549683 | 0.523926 | 18 | 34 | 0.654631 | 0.700123 | 18 |

**Table** 4 shows the numerical results achieved by the proposed JBUA algorithm for solving above problem. 34 randomly generated initial relaxation factors were used in this experiment. First observation is that whatever are the initial relaxation factors, the proposed algorithm adapts the relaxation factors dynamically and the algorithm converges rapidly. Another observation is that the proposed algorithm is less sensitive to the relaxation factors.

**Table 5**
Number of iteration required by the classical Jacobi-SR method for various randomly generated relaxation factors.

| Value of $\omega$ | 0.00212 | 0.200000 | 0.500 | 0.7 | 0.75 | 0.79 | 0.80 |
|---|---|---|---|---|---|---|---|
| Iteration No. | 100000000 | 125 | 41 | 25 | 22 | 19 | 19 |
| | | | | | | | |
| Value of $\omega$ | **0.81\*** | **0.815\*** | 0.85 | 0.9 | 1.0 | 1.5 | 1.6 |
| Iteration No. | **18** | **18** | 22 | 26 | 37 | Diverge | Diverge |

**Table** 5 represents the numerical results achieved by the classical Jacobi-SR method for solving above problem with several randomly generated initial relaxation factors. From this experiment it is observed that classical Jacobi-SR method is very much sensitive to the relaxation factor. Also notice that the optimal relaxation $\omega^*$ for classical Jacobi-SR is very near to 0.810 and required iteration number, by using classical Jacobi-SR method, is 18.

**Table 6**
Comparison between existing Classical SR (both for Jacobi and Gauss-Seidel based) techniques and proposed JBUA hybrid algorithms for several randomly generated test problems

| Domain of the elements of the coefficient matrix **A** & right side constant vector **b** of some test Problems | Relaxation factor $\omega$ | Iteration (for Jacobi-SR) | Iteration Gauss-Sedel-SR) | Relaxation factors $\omega_1$ $\omega_2$ | | Generation (For proposed JBUA) |
|---|---|---|---|---|---|---|
| $a_{ii} = \begin{cases} 3n & i \leq 50 \\ 4n & i > 50 \end{cases}$ & $b_i = i$; for $i = 1, 2, \cdots, n$; $a_{ij} = (-1)^i j$; for $i \neq j$ & $i, j = 1, 2, \cdots, n$ | 0.01 | 3529 | 3531 | 0.01 | 1.9 | 299 |
| | 0.05 | 1704 | 1583 | 0.1 | 1.8 | 304 |
| | 0.1 | 702 | 400 | 0.05 | 1.5 | 301 |
| | 1.3 | 1264 | 975 | 0.1 | 1.6 | 290 |
| | 1.5, 1.6 | Diverge | Diverge | 0.2 | 1.4 | 310 |
| | 1.8, 19 | Diverge | Diverge | 0.3 | 1.9 | 311 |
| | 1.7 | Diverge | Diverge | 0.3 | 1.5 | 307 |
| | 1.9 | Diverge | Diverge | 0.11 | 1.2 | 297 |
| $a_{ii} = \begin{cases} -3n & i \leq 50 \\ 4n & i > 50 \end{cases}$ and $b_i = i$; for $i = 1, 2, \cdots, n$; $a_{ij} = j$; for $i \neq j$ & $i, j = 1, 2, \cdots, n$ | 0.001 | 5789 | 5467 | 0.01 | 1.3 | 103 |
| | 0.01 | 3621 | 3590 | 0.1 | 1.9 | 107 |
| | 0.1 | 349 | 346 | 0.001 | 0.1 | 138 |
| | 1.0 | Diverge | 110 | 0.01 | 1.8 | 107 |
| | 1.5, 1.6 | Diverge | Diverge | 1.6 | 1.9 | 106 |
| | 1.7, 1.8 | Diverge | Diverge | 0.5 | 1.6 | 103 |
| $a_{ii} = (-1)^i n \cdot i$ & $b_i = i$; for $i = 1, 2, \cdots, n$; $a_{ij} = j$; for $i \neq j$ & $i, j = 1, 2, \cdots, n$ | 0.01 | 4321 | 4123 | 0.01 | 1.3 | 17 |
| | 0.1 | 348 | 348 | 0.1 | 1.9 | 18 |
| | 0.3 | 104 | 104 | 0.2 | 0.9 | 17 |
| | 1.5 | 59 | 40 | 0.01 | 1.8 | 21 |
| | 1.7 | 121 | 102 | 0.3 | 1.4 | 19 |
| | 1.9 | 490 | 425 | 0.01 | 1.3 | 18 |
| $a_{ii} = -n \cdot i$ & $b_i = 2n \cdot i$; for $i = 1, 2, \cdots, n$; $a_{ij} = \pm 4j$; for $i \neq j$ & $i, j = 1, 2, \cdots, n$ | 0.01 | 17000 | 17500 | 0.01 | 1.9 | 38 |
| | 0.1 | 14900 | 15000 | 0.1 | 0.5 | 37 |
| | 0.2 | 165 | 175 | 0.2 | 1.2 | 36 |
| | 1.5 | 133 | 20000 | 0.01 | 1.8 | 37 |
| | 1.75 | 1264 | 25000 | 0.3 | 1.4 | 40 |
| | 1.9 | Diverge | Diverge | 1.0 | 1.5 | 36 |

*Note that the threshold error for all the cases is $\eta = 10^{-08}$

**Table 6** represents further numerical results achieved by the classical SR technique (both for Jacobi and Gaiss-Seiodel based) and proposed JBUA algorithm for solving above-mentioned problems (See 1st column of the Table 6) with several randomly selected initial relaxation factors. First observation is that whatever are the initial relaxation factors, the proposed JBUA algorithm adapts the relaxation factors dynamically and the algorithm converges rapidly. Another observation is that the proposed algorithm is very less sensitive to the relaxation factors and both classical Jacobi-SR method and Gauss-Seidel-SR method are very much sensitive to the relaxation factors. This indicates that the simple adaptation scheme for relaxation factors had worked quite effectively in the proposed JBUA hybrid algorithm. From this experiment it is observed that

**Table 7**
Comparison between existing GSBUA and proposed JBUA hybrid algorithms for several randomly generated test problems

| Label of Test Function | Domain of the elements of the coefficient matrix **A** & right side constant vector **b** of test Problems | | | Threshold Error, η | GSBUA Hybrid Algorithm | Proposed JBUA Hybrid Algorithm |
|---|---|---|---|---|---|---|
| | | | | | Number of Generation | Number of Generation |
| $P_1$ | $a_{ii} = 70$; | $a_{ij} \in (-10,10)$, | $b_i \in (-70,70)$ | $10^{-12}$ | 166 | 190 |
| $P_2$ | $a_{ii} \in (50,100)$; | $a_{ij} \in (-10,10)$ ; | $b_i \in (-100,100)$ | $10^{-12}$ | 85 | 91 |
| $P_3$ | $a_{ii} \in (1,100)$; | $a_{ij} \in (-2,2)$; | $b_i = 2$ | $10^{-12}$ | 559 | 586 |
| $P_4$ | $a_{ii} = 200$; | $a_{ij} \in (-30,30)$; | $b_i \in (-400,400)$ | $10^{-11}$ | 156 | 175 |
| $P_5$ | $a_{ii} \in (-70,70)$; | $a_{ij} \in (0,4)$ ; | $b_i \in (0,70)$ | $10^{-08}$ | 801 | 816 |
| $P_6$ | $a_{ii} \in (-200,200)$; | $a_{ij} \in (-10,10)$; | $b_i \in (-100,100)$ | $10^{-11}$ | 189 | 200 |
| $P_7$ | $a_{ii} \in (-100,100)$; | $a_{ij} \in (-10,10)$; | $b_i \in (-200,200)$ | $10^{-06}$ | 5683 | 5711 |
| $P_8$ | $a_{ii} \in (10,50)$; | $a_{ij} \in (5,8)$; | $b_i \in (-200,200)$ | $10^{-11}$ | 618 | 655 |
| $P_9$ | $a_{ii} \in (100,300)$; | $a_{ij} \in (-50,50)$; | $b_i \in (-100,100)$ | $10^{-11}$ | 1508 | 1832 |
| $P_{10}$ | $a_{ii} \in (200,300)$; | $a_{ij} \in (-100,100)$; | $b_i \in (-100,100)$ | $10^{-11}$ | 798 | 870 |

To evaluate the proposed JBUA hybrid algorithm further, ten test problems, labeled from $P_1$ to $P_{10}$, with dimension, $n=100$ were considered. For each test problem $P_i : i =1, 2, \ldots 10$, the elements of the coefficient matrix **A** and elements of the constant vector **b** were all generated uniformly and randomly within given boundaries (shown in 2nd column with corresponding rows of **Table** 7 ). But each coefficient matrix **A**, constant vector **b** and initial population **X** were identical for each comparison. Initial relaxation factors are set at $\omega_1 = 0.5$ and $\omega_2 = 1.5$ for all the cases. For different problems $P_1 − P_{10}$ different threshold of errors, η, were allowed. **Table** 7 shows the comparison of the number of generation (iteration) needed by the GSBUA hybrid algorithm and that needed by the proposed JBUA hybrid algorithm to solve the linear equations Eq. (2) to the given preciseness, $\eta$ (see column three of **Table** 7). One observation can be made immediately from the **Table** 7 that, the numbers of generations are comparable in each case in the both hybrid algorithms (proposed JBUA and existing GSBUA algorithms) in sequential computing

## 6. Parallel Processing

Parallel searching is one of the main properties of Evolutionary Computational (EC) techniques. As computers can be used for parallel searching by using parallel processors, so EC techniques can be used to solve various kinds of complex problems. For available of parallel processors, recently, evolutionary algorithms are well developed and successfully used to solve so many real world problems. Though individuals of population can be implemented in parallel processing environment for both GSBUA hybrid algorithm and JBUA hybrid algorithm. But inherently, GSBUA hybrid algorithm cannot be implemented in parallel processing environment efficiently. Because classical Gauss-Seidel method, cannot be implemented in parallel processing environment efficiently. Whereas, inherently proposed JBUA hybrid algorithm can be implemented in parallel processing environment efficiently.. Since Jacobi-SR method can be implemented in parallel processing environment and efficiency of this method is near to one [1]. As a result by using parallel processors it can be reduced a large amount of times for each iteration of JBUA algorithm. For example, if $n^2$ processors are available and using these processors inherently, then JBUA hybrid algorithm reduces the time for each iteration to $\log_2 n$ time units. This is a significant speedup over the sequential algorithm that requires $n^2$ time units per iteration (as GSBUA hybrid algorithm inherently do) [1].

## 7. Concluding Remarks

In this paper, Jacobi-SR based hybrid evolutionary algorithm has been proposed for solving linear systems of equations. The proposed Jacobi-based hybrid algorithm integrates the classical Jacobi-SR method with evolutionary computation techniques. The recombination operator in the algorithm mixed all parents with a probabilistic ratio by a stochastic matrix, which is similar to the intermediate recombination often used in evolution strategies [8,9]. The mutation operator is equivalent to one iteration in the classical Jacobi-SR method.

However, the mutation is stochastic as a result of stochastic self-adaptation of the relaxation factor $\omega$. Individuals of population can be used in parallel processing environment for both Gauss-Seidel-SR based hybrid algorithm and Jacobi-SR based hybrid algorithm. But as Gauss-Seidel method, inherently, cannot be matched with parallel processing environment, so Gauss-Seidel based hybrid algorithm, inherently, cannot be implemented in parallel computing environment efficiently. On the other hand, as Jacobi-SR algorithm is matched with parallel processing environment efficiently, so proposed JBUA hybrid algorithm, inherently, can be implemented in parallel processing environment efficiently. As a result by using parallel computing environment, the proposed JBUA algorithm reduces a significant amount of time. For example, if $n^2$ processors are available, then proposed algorithm reduces to $log_2 n$ time units for each generation. This is a significant speedup over the sequential algorithm, which requires $n^2$ time units per iteration [1].

These primary numerical experiments with various test problems have shown that the proposed Jacobi-SR based hybrid evolutionary algorithm performs much better than the Gauss-based hybrid algorithm as well as both classical Jacobi-SR method and Gauss-Seidel-SR method. The proposed hybrid algorithm is efficient and robust than both classical Jacobi-SR method and Gauss-Seidel-SR method. And it is converged very rapidly compare to both these classical methods. The proposed algorithm is less sensitive to the relaxation factors. Whereas both classical Jacobi and Gauss-Seidel based SR methods are very much sensitive to the relaxation factors. The proposed hybrid algorithm is also very simple and easy to implement both in sequential and parallel computing environment.